\documentclass[]{bytedance}
\usepackage[toc,page,header]{appendix}

\usepackage{lineno} 
\usepackage{minitoc}
\usepackage{amsfonts}
\usepackage{amssymb}
\usepackage{tabularx}
\usepackage{listings}
\usepackage{xcolor}
\usepackage{cancel}

\usepackage{tabulary,multirow,xspace}
\usepackage{mathtools,nicefrac,mmstyle}
\usepackage{subcaption}
\usepackage{caption}
\usepackage{float}
\usepackage{wrapfig} 
\usepackage{colortbl}

\usepackage{multicol}
\usepackage[most]{tcolorbox}

\usepackage{pifont} 

\definecolor{codegreen}{rgb}{0,0.6,0}
\definecolor{codegray}{rgb}{0.5,0.5,0.5}
\definecolor{codepurple}{rgb}{0.58,0,0.82}
\definecolor{backcolour}{rgb}{0.95,0.95,0.92}
\definecolor{boxblue}{RGB}{57,89,163}
\definecolor{boxbluebg}{RGB}{230,237,250} 
\definecolor{commentgreen}{rgb}{0.1, 0.4, 0.1}
\definecolor{keywordblue}{rgb}{0.1, 0.1, 0.7}
\definecolor{stringred}{rgb}{0.7, 0.1, 0.1}

\lstdefinestyle{mystyle}{
    backgroundcolor=\color{backcolour},   
    commentstyle=\color{commentgreen},
    keywordstyle=\color{keywordblue},
    stringstyle=\color{stringred},
    numberstyle=\tiny\color{codegray},
    basicstyle=\ttfamily\scriptsize, % Changed to scriptsize to match your second definition
    breakatwhitespace=false,         
    breaklines=true,                 
    captionpos=b,                    
    keepspaces=true,                 
    numbers=none,                    
    numbersep=5pt,                  
    showspaces=false,                
    showstringspaces=false,
    showtabs=false,                  
    tabsize=2,
    frame=none,                     
    language=Python
}
\definecolor{mygray1}{gray}{.95}
\definecolor{mygray2}{gray}{.9}
\definecolor{mygray3}{gray}{.95}

\newlength\savewidth
\newcolumntype{x}[1]{>{\centering\arraybackslash}p{#1pt}}

\newcommand{\app}{\raise.17ex\hbox{$\scriptstyle\sim$}}

\definecolor{tech-green}{HTML}{00A388} 

\definecolor{tech-blue}{HTML}{0077CC}

\definecolor{tech-purple}{HTML}{9B51E0}

\title{
    {
     \textcolor{tech-green}{\bfseries Jo}\textcolor{tech-blue}{\bfseries V}\textcolor{tech-purple}{\bfseries A}:} {
     Unified Multimodal Learning for \\
     \textcolor{tech-green}{\bfseries Jo}int
     \textcolor{tech-blue}{\bfseries V}ideo-\textcolor{tech-purple}{\bfseries A}udio Generation and Editing
    }
}

\author{
    Xiaohu Huang\textsuperscript{1, *} \qquad
    Haoyang He\textsuperscript{2, *} \qquad
    Hao Zhou\textsuperscript{3, *} \\
    Qiangpeng Yang\textsuperscript{3} \qquad
    Shilei Wen\textsuperscript{3} \qquad
    Kai Han\textsuperscript{1, $\dagger$} \\[1em]
    \textsuperscript{1}The University of Hong Kong \qquad
    \textsuperscript{2}Zhejiang University \qquad
    \textsuperscript{3}ByteDance \\
    \small
    \textsuperscript{*}Equal contribution \qquad
    \textsuperscript{$\dagger$}Corresponding author
}

\abstract{
In this paper, we present {\textcolor{tech-green}{\bfseries Jo}\textcolor{tech-blue}{\bfseries V}\textcolor{tech-purple}{\bfseries A}}, a streamlined framework that unifies joint video-audio generation and editing. While existing methods often rely on fragmented, task-specific architectures or complex fusion mechanisms, JoVA employs native joint representation learning for direct video, audio, and text interaction in a dual-branch architecture. This design eliminates redundant alignment modules and effectively unifies diverse multimodal tasks within a single model. Furthermore, we utilize channel-wise conditioning for flexible image and video reference to avoid massive token expansion, alongside a mouth-area loss to enhance lip alignment. To fully empower and systematically evaluate this framework, we construct a comprehensive training corpus encompassing video-audio generation and editing datasets, and introduce unified benchmarks tailored for these multimodal tasks. Extensive experiments demonstrate that JoVA achieves state-of-the-art performance across benchmarks, establishing it as an extensible framework for versatile content creation. Project page: \url{https://visual-ai.github.io/jova}
}

\correspondence{Kai Han at \email{kaihanx@hku.hk}, Xiaohu Huang at \email{huangxiaohu@connect.hku.hk}}

\begin{document}
\maketitle
\section{Introduction}

Recent years have witnessed a remarkable evolution in AI-driven content creation, transitioning from image synthesis to the more complex domain of video generation. This surge has catalyzed extensive research across various specialized directions. For instance, video generation models~\cite{wang2025wan,Hunyuanvideo,yang2024cogvideox,Stable_video_diffusion,Cogvideo} have achieved high-fidelity visual synthesis; joint video-audio generation frameworks~\cite{UniVerse,ltx2,Ovi} have emerged to produce synchronized content; video editing techniques~\cite{lucyedit,icve,ditto} enable fine-grained visual modifications; and lip-sync methods~\cite{wav2lip,mukhopadhyay2024diff2lip,li2024latentsync} focus on simulating realistic talking heads. However, these directions largely exist as separate streams of work.

To tackle these distinct tasks, researchers typically design task-specific model architectures or introduce complex multimodal fusion mechanisms, such as additional cross-attention layers~\cite{Ovi,UniVerse,ltx2}. This fragmentation limits model simplicity and extensibility. Recently, unified models~\cite{wei2025univideo,omnivideo,jiang2025vace} have demonstrated the ability to handle multiple video tasks within a single framework. Yet, they lack the capability to output audio, rendering the generated video content less vivid and immersive. Therefore, developing a unified model capable of seamlessly handling diverse multimodal tasks---spanning both generation and editing across video and audio modalities---is a highly promising yet challenging frontier.

\begin{figure}[t]
\centering
\includegraphics[width=0.94\linewidth]{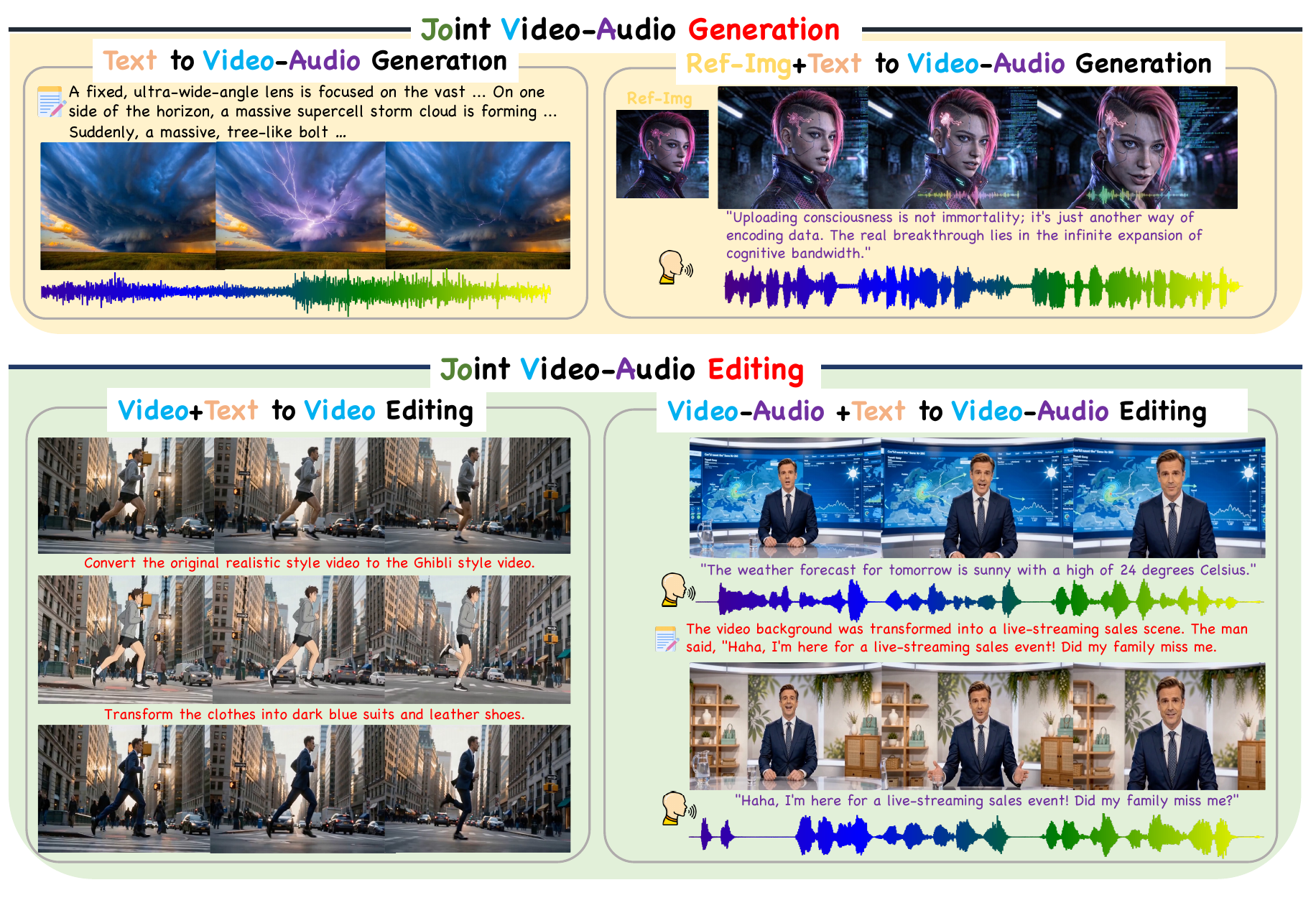}
\caption{
\textbf{Overview of the JoVA framework's capabilities.} JoVA seamlessly unifies various multimodal tasks within a single model, including text-to-video-audio generation, reference-image-conditioned generation, video editing, and joint video-audio editing. By utilizing a simplified architecture, it ensures high-quality, synchronized, and flexible multimodal content creation.
}
\label{fig:teaser}
\end{figure}

To bridge this gap, we present JoVA, a unified framework capable of joint video-audio generation and editing within a single model (Fig.~\ref{fig:teaser}). To realize this architecture, we integrate pretrained video and audio models, streamlining the overall design by eliminating redundant multimodal fusion modules. In their place, we use a native joint self-attention mechanism that concatenates video, audio, and text tokens, enabling direct and seamless multimodal interaction. This simplified configuration preserves the core philosophy of transformer architectures and guarantees exceptional task extensibility. To handle visual reference conditions in editing tasks, we adopt a channel-wise conditioning approach, which flexibly incorporates spatial inputs without drastically expanding the token sequence length. Furthermore, to address the challenge of lip-speech synchronization---where the mouth region occupies a tiny fraction of the frame (\(\sim\)2.3\%)---we incorporate an area-specific loss during training to ensure precise alignment without adding architectural complexity.

To endow the model with versatile capabilities, we construct a large-scale training corpus of \(\sim\)3M diverse pairs, encompassing natural video-audio scenes, human talking videos, and complex joint editing data. For assessment, we introduce two benchmarks, JoVABench-Gen and JoVABench-Edit, and further evaluate our model on the public Verse-Bench~\cite{UniVerse}. Extensive experiments demonstrate that our model consistently achieves leading performance across all tasks. Please refer to the supplementary material for the generated video results.

In summary, our main contributions are as follows:
\begin{itemize}
\item We propose JoVA, a unified architecture for joint video-audio modeling. By processing modalities through a unified representation, JoVA eliminates the reliance on complex cross-modal fusion modules and fragmented designs, establishing a highly extensible foundation for multimodal content creation.
\item To empower this single architecture to handle a diverse spectrum of generation and editing scenarios, we construct a multi-domain training corpus of \(\sim\)3M high-quality pairs. The introduced automated data synthesis pipeline is designed to scale up the generation and editing of data.
\item Extensive experiments on two introduced benchmarks (JoVABench-Gen and JoVABench-Edit) and Verse-Bench demonstrate that our single JoVA model achieves superior performance across diverse generation and editing tasks.
\end{itemize}
\section{Related Work}

\subsection{Joint Video-Audio Generation}

Recent advances in video generation have been driven by diffusion models transitioning from UNet-based architectures~\cite{unet,rombach2022ldm} to transformer-based designs~\cite{dit}, exemplified by Sora~\cite{sora,sora2_website}. This shift has spurred high-fidelity models like Wan~\cite{wang2025wan,wan25_website}, Goku~\cite{chen2025goku}, Waver~\cite{zhang2025waver}, CogVideoX~\cite{yang2024cogvideox}, and HunyuanVideo~\cite{Hunyuanvideo}. Beyond visual fidelity, recent post-training methods such as VGGRPO~\cite{an2026vggrpo} introduce latent 4D geometry rewards to improve camera stability and world consistency in generated videos. Building upon this visual foundation, cross-modal generation largely diverged into unidirectional streams. Audio-driven video methods~\cite{chen2025hunyuanavatar,OmniAvatar,Wan-s2v,lin2025omnihuman} typically rely on cascaded pipelines to animate talking heads from speech. Conversely, video-to-audio (V2A) approaches~\cite{MMAudio,avlink,Hunyuanvideo-foley,Diff-foley,Foleycrafter,kling-foley} focus on generating ambient sounds for silent videos. However, these decoupled approaches often require auxiliary synchronization modules, increasing system complexity and limiting semantic coherence. To achieve cohesive multimodal synthesis, various joint video-audio generation frameworks~\cite{ishii2024simple,seehear,AV-DiT,javisdit,mmdifusion,zhao2025uniform} have emerged. Despite the significant progress marked by large-scale models like UniVerse-1~\cite{UniVerse}, Ovi~\cite{Ovi}, UniAVGen~\cite{zhang2025uniavgen}, and LTX-2~\cite{ltx2}, most existing approaches adopt dual-branch architectures that process modalities separately and fuse them through explicit cross-attention. Such newly introduced alignment layers are typically trained on much smaller multimodal corpora than the billion-scale video backbones, which can disturb pretrained visual priors. In contrast, JoVA abandons complex fusion modules for native joint self-attention, reusing pretrained transformer parameters to process all modalities equally for precise alignment and architectural elegance.

\subsection{Joint Video-Audio Editing}

Instruction-driven video editing has advanced rapidly, with existing methods injecting conditions via cross-attention~\cite{omnivideo}, channel or sequence concatenation~\cite{lucyedit,wei2025univideo,icve,he2025openve}, or parameter-efficient fine-tuning~\cite{mou2025instructx}. Yet handling complex multi-task instructions in a single model remains difficult, and preserving audio-video synchronization adds further complexity. Recent works~\cite{Object-AVEdit,aved,AVI-Edit} employ training-free latent inversion or intricate pipelines with mask refiners and feedback agents. Similarly, precise lip-sync methods~\cite{mukhopadhyay2024diff2lip,wav2lip,li2024latentsync} are typically restricted to editing only the mouth region and heavily depend on external masks for guidance. A major drawback of these approaches is their fragmented nature: they often require stitching together independent models and auxiliary inputs, severely complicating the pipeline. In contrast, JoVA resolves these limitations through a natively unified architecture. It utilizes joint self-attention for multimodal interaction and channel-wise conditioning to avoid drastic sequence expansion, while a simple area-specific loss ensures accurate lip-sync.

\section{Method}

\subsection{Preliminary}

Our framework builds upon the Waver~\cite{zhang2025waver} video generation model, which employs a transformer-based diffusion architecture. For video encoding, we adopt the Wan2.1 VAE~\cite{wang2025wan} to compress visual features in both spatial and temporal dimensions. Text prompts and instructions are processed through a combination of T5-XXL~\cite{raffel2020t5} and Qwen2.5-32B~\cite{qwen2.5} encoders to provide rich semantic conditioning.

Following the MM-DiT architecture~\cite{mmdit}, Waver processes multimodal inputs through a two-stage design: an initial multi-stream stage where modalities maintain separate parameters, followed by a single-stream stage where all modalities share the same parameters. Each stage consists of transformer blocks containing self-attention and feed-forward network (FFN) modules. For a given layer, the computation can be expressed as:
\begin{equation}
\mathbf{h}' = \text{SelfAttn}(\mathbf{h}) + \mathbf{h}, \quad \mathbf{h}'' = \text{FFN}(\mathbf{h}') + \mathbf{h}',
\end{equation}
where \(\mathbf{h} \in \mathbb{R}^{N \times D}\) denotes the input features with \(N\) tokens and dimension \(D\).

For training, we employ the flow matching~\cite{flowmatching} framework, which provides a simple and effective approach for generative modeling. The training objective minimizes the difference between the predicted velocity field and the true flow velocity:
\begin{equation}
\mathcal{L}_{\text{video}} = \mathbb{E}_{t, \mathbf{x}_0, \mathbf{x}_1, \mathbf{C}} \left\| v_\theta(t, \mathbf{C}, \mathbf{x}_t) - u(\mathbf{x}_t | \mathbf{x}_0, \mathbf{x}_1) \right\|^2,
\end{equation}
where \(t \sim \mathcal{U}[0,1]\), \(\mathbf{x}_0 \sim \mathcal{N}(0, \mathbf{I})\) is sampled noise, \(\mathbf{x}_1 \sim q(\mathbf{x}_1, \mathbf{C})\) represents training data conditioned on \(\mathbf{C}\) (text and other modalities), \(\mathbf{x}_t = t\mathbf{x}_1 + (1-t)\mathbf{x}_0\) defines the interpolation path, and \(u(\mathbf{x}_t | \mathbf{x}_0, \mathbf{x}_1) = \mathbf{x}_1 - \mathbf{x}_0\) denotes the corresponding flow velocity.

\begin{figure*}[t]
\centering
\includegraphics[width=0.93\linewidth]{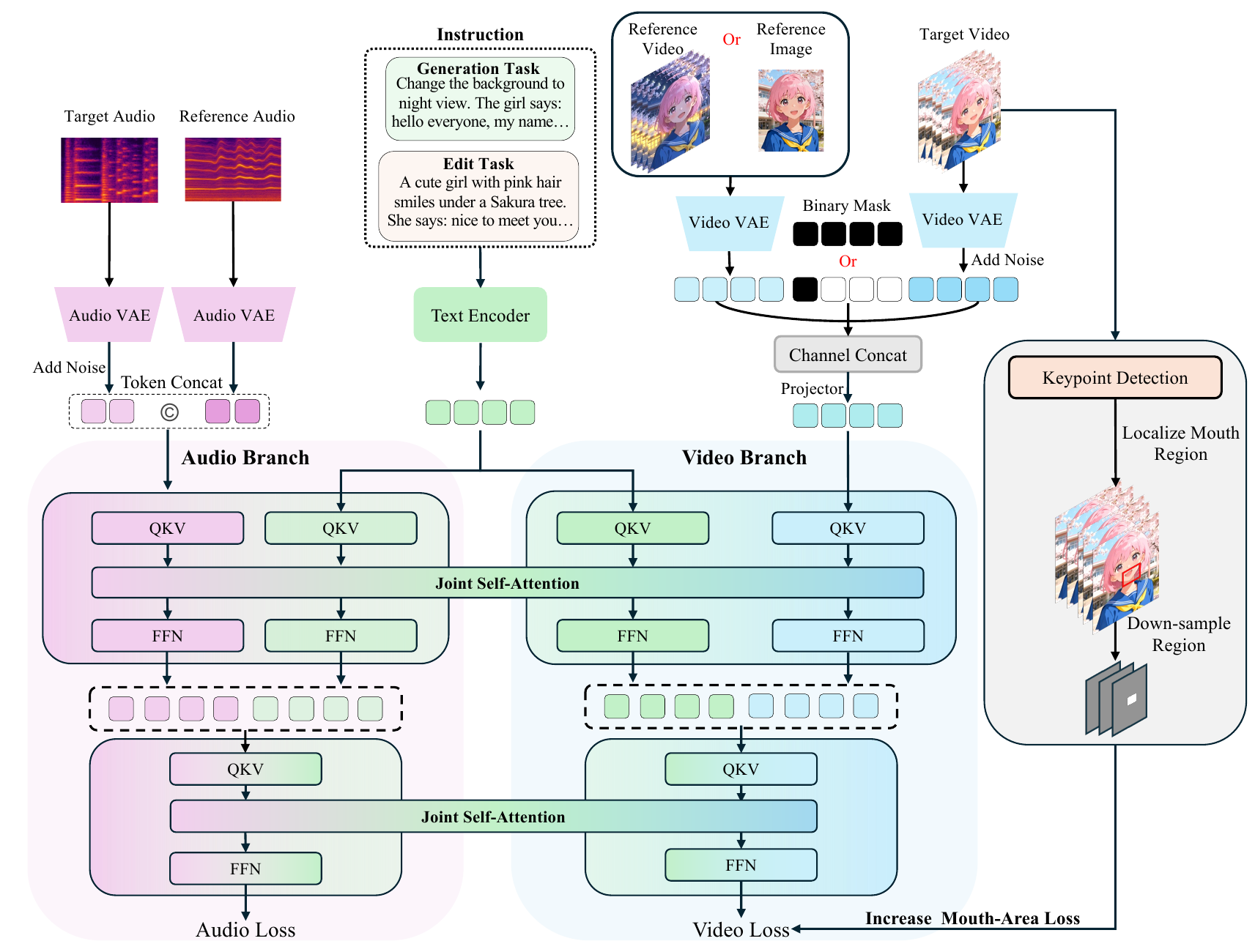}
\caption{
\textbf{Overall architecture of the proposed JoVA framework.} Our model natively unifies video and audio generation and editing through a joint self-attention mechanism. Complex multi-task instructions are processed via a text encoder to guide the generation. For visual conditioning, we employ a channel-wise concatenation strategy with a binary mask to flexibly accommodate reference videos or images. Conversely, reference audio tokens are encoded and integrated into the audio branch by token concatenation. A localized mouth-area loss is applied to the video output to ensure lip-speech synchronization during training, which is exclusively active for human speech scenarios and does not affect the generation of natural scenes or ambient sounds.
}
\label{fig:method}
\end{figure*}

\subsection{Unified Video-Audio Generation and Editing}

\noindent \textbf{Joint Video-Audio Architecture}
We propose a unified framework designed to natively handle both generation and editing tasks across video and audio modalities. As illustrated in Fig.~\ref{fig:method}, our architecture comprises a visual branch, an audio branch, and a shared text encoder that processes complex multi-task instructions. Instead of relying on cascaded models or complex external fusion networks, we seamlessly integrate these modalities into a shared transformer backbone.

To establish robust audio generation capabilities within this single model, we initialize the audio diffusion model by duplicating the pretrained parameters of our video backbone. For feature extraction, we employ the MMAudio VAE~\cite{MMAudio} to process audio spectrograms, converting raw audio signals into compact latent representations. Within each transformer block, tokens from the video input, the audio input, and the instruction text are processed through a shared joint self-attention mechanism:
\begin{equation}
[\mathbf{h}'_v; \mathbf{h}'_a; \mathbf{h}'_{t}] = \text{JointAttn}([\mathbf{h}_v; \mathbf{h}_a; \mathbf{h}_{t}]),
\end{equation}
where \(\mathbf{h}_{t}\) contains text features for both the video and audio branches. Unlike prior methods that rely on separate cross-attention layers, our joint self-attention mechanism allows video, audio, and text tokens to interact directly in a unified space. It also avoids adding newly initialized alignment modules whose training data scale is much smaller than that of the pretrained backbone. Furthermore, to ensure precise temporal alignment between the video and audio streams, we adopt the temporal-aligned RoPE (Rotary Position Embedding) from MMAudio~\cite{MMAudio}, which synchronizes the positional encodings of both modalities along the temporal dimension.

\noindent \textbf{Unified Conditioning.} To accommodate editing instructions, we introduce modality-specific conditional inputs. The integration strategies are carefully tailored to the distinct sequence characteristics of video and audio modalities.

For the audio branch, alongside the noisy target audio latent, we introduce a reference audio latent encoded via the Audio VAE to provide acoustic context. Because the sequence length of audio latents is extremely compact---typically accounting for less than \(1\%\) of the video token length---we can directly employ a token concatenation strategy to integrate the reference audio. This approach effectively provides the necessary context without incurring significant computational overhead.

Conversely, for the visual branch, directly appending reference video frames via token concatenation would lead to a massive increase in sequence length, resulting in prohibitive computational costs. Therefore, we employ a channel-wise conditioning strategy guided by a binary mask. Specifically, the reference visual input is encoded by the Video VAE and concatenated with the noisy target video latent and a corresponding mask along the channel dimension. The mask is set to \(1\) for frames where the reference is provided, and \(0\) otherwise (in which case the reference channels are filled with default padding). After the channels are concatenated, a projector layer is utilized to fuse their information. This design not only avoids the token explosion problem but also natively unifies image-to-video generation and video editing (flexibly accommodating reference images as conditions) without requiring architectural modifications.

\noindent \textbf{Training Objectives and Mouth-Area Loss}
Our unified model is optimized using flow matching objectives. To ensure the model learns a wide variety of acoustic patterns and human speech characteristics, the audio branch is trained on a comprehensive mixture of large-scale text-to-audio and text-to-speech datasets. The audio flow matching loss is defined as:
\begin{equation}
\mathcal{L}_{\text{audio}} = \mathbb{E}_{t, \mathbf{a}_0, \mathbf{a}_1, \mathbf{C}} \left\| v_\theta(t, \mathbf{C}, \mathbf{a}_t) - u(\mathbf{a}_t | \mathbf{a}_0, \mathbf{a}_1) \right\|^2,
\end{equation}
where \(\mathbf{a}_t\) represents the audio latent at time \(t\), and \(\mathbf{C}\) denotes the conditioning information.

While the unified architecture effectively handles general video-audio alignment, achieving fine-grained lip-speech synchronization for human-centric content requires targeted supervision. To address this, we introduce a localized strategy that emphasizes the mouth region during training. We first apply facial keypoint detection to the original video frames to localize the mouth region and extract a bounding box. This bounding box is then mapped to the VAE latent space through proportional spatial scaling and a sliding-window temporal downsampling to match the latent dimensions, yielding a precise mouth region mask \(\mathcal{M}\). More implementation details are given in the supplementary material.

Our final training objective combines the general flow matching losses with this localized supervision:
\begin{equation}
\mathcal{L}_{\text{total}} = \mathcal{L}_{\text{video}} + \mathcal{L}_{\text{audio}} + \lambda \mathcal{L}_{\text{mouth}},
\label{eq:total_loss}
\end{equation}
where \(\mathcal{L}_{\text{video}}\) is the standard video flow matching loss. The term \(\mathcal{L}_{\text{mouth}}\) applies the flow matching objective strictly restricted to the mouth region of the video latent:
\begin{equation}
\mathcal{L}_{\text{mouth}} = \mathbb{E}_{t, \mathbf{v}_0, \mathbf{v}_1, \mathbf{C}} \left\| v_\theta(t, \mathbf{C}, \mathbf{v}_t)_{\mathcal{M}} - u(\mathbf{v}_t | \mathbf{v}_0, \mathbf{v}_1)_{\mathcal{M}} \right\|^2,
\end{equation}
where \(\mathbf{v}_t\) denotes the video latent at time \(t\), and \(\lambda\) is a weighting coefficient. Crucially, this mouth-area loss is exclusively applied to data containing human speech. For natural scenes or general ambient sounds, we set \(\lambda\) to \(0\), ensuring that the generation of non-human content remains completely unaffected.

\subsection{Construction of Training Data}

\begin{figure*}[t]
\centering
\includegraphics[width=0.93\linewidth]{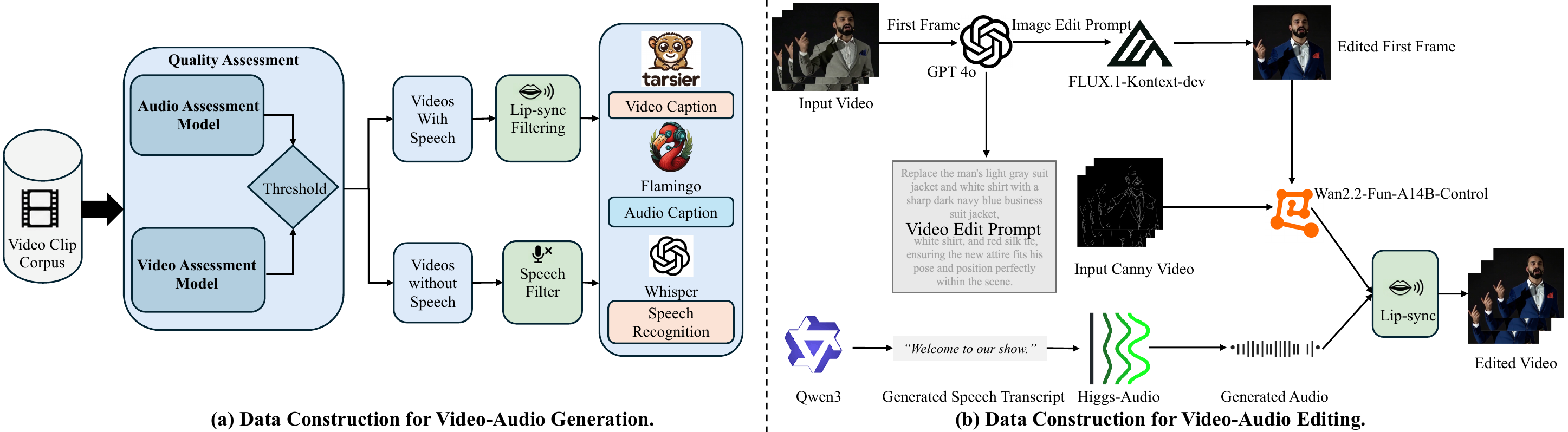}
\caption{
\textbf{Overview of our training data construction pipelines.}
(a) \textbf{Video-Audio Generation Data:} Raw videos undergo rigorous quality assessment and filtering, followed by comprehensive multimodal annotation including video/audio captioning and speech recognition.
(b) \textbf{Video-Audio Editing Data:} A workflow to synthesize paired editing data, utilizing foundation models for visual editing, audio generation, and lip-sync alignment.
}
\label{fig:dataset}
\end{figure*}

To support unified video-audio generation and editing across diverse scenarios, we construct a large-scale, high-quality dataset comprising approximately 3 million (3M) video-audio-text pairs. Fig.~\ref{fig:dataset_distribution} summarizes the dataset composition, while Fig.~\ref{fig:dataset} illustrates the two primary construction pipelines tailored for generation and editing tasks, respectively.

\noindent \textbf{Data Construction for Video-Audio Generation.} As shown in Fig.~\ref{fig:dataset}(a), we begin with a large corpus of raw video clips. To ensure data quality, these clips first pass through a rigorous Quality Assessment module, utilizing both Audio and Video Assessment Models to filter out low-quality samples based on predefined thresholds. The retained high-quality videos are then categorized into two streams: videos with speech and videos without speech. For videos containing speech, we apply a Lip-sync Filtering module to ensure tight audio-visual alignment, which is crucial for talking-head generation. For videos without speech, a Speech Filter is applied to guarantee the purity of ambient sounds. Subsequently, we employ a multimodal annotation pipeline to generate comprehensive text labels. We utilize Tarsier~\cite{yuan2025tarsier2} for detailed video captioning and Audio-flamingo~\cite{goel2025audioflamingo3} for audio captioning. For the speech subset, Whisper~\cite{whisper} is additionally used for automatic speech recognition (ASR) to extract precise transcripts. This tri-level annotation enriches each sample with structured semantic information from complementary visual, auditory, and linguistic perspectives.

\noindent \textbf{Data Construction for Video-Audio Editing.} To train the model for video-audio editing tasks, we design an automated data synthesis pipeline (Fig.~\ref{fig:dataset}b) to construct paired source-target editing data. Inspired by the visual editing paradigm of OpenVE~\cite{he2025openve}, we adapt and extend this approach to the multimodal domain. Given an input video, we extract its first frame and utilize GPT-4o to generate an Image Edit Prompt. This prompt guides FLUX.1-Kontext-dev~\cite{labs2025flux} to produce an Edited First Frame. Concurrently, GPT-4o~\cite{gpt4o} generates a Video Edit Prompt detailing the desired temporal changes. We then extract the Canny edge map from the input video to serve as structural guidance. The Edited First Frame, Input Canny Video, and Video Edit Prompt are fed into Wan2.2-Fun-A14B-Control~\cite{wan2_fun} to generate the visual component of the edited video. Building upon this visual foundation, we further construct the corresponding audio track. We use Qwen3~\cite{yang2025qwen3} to generate a speech transcript, which is then synthesized into high-quality audio using Higgs-Audio~\cite{boson2025higgs}. Finally, a Lip-sync module aligns the generated audio with the visual output, resulting in the final Edited Video. This pipeline allows us to efficiently scale up high-quality, paired multimodal editing data.

\subsection{Dataset Distribution and Details}

\begin{figure*}[t]
    \centering
    \includegraphics[width=0.95\linewidth]{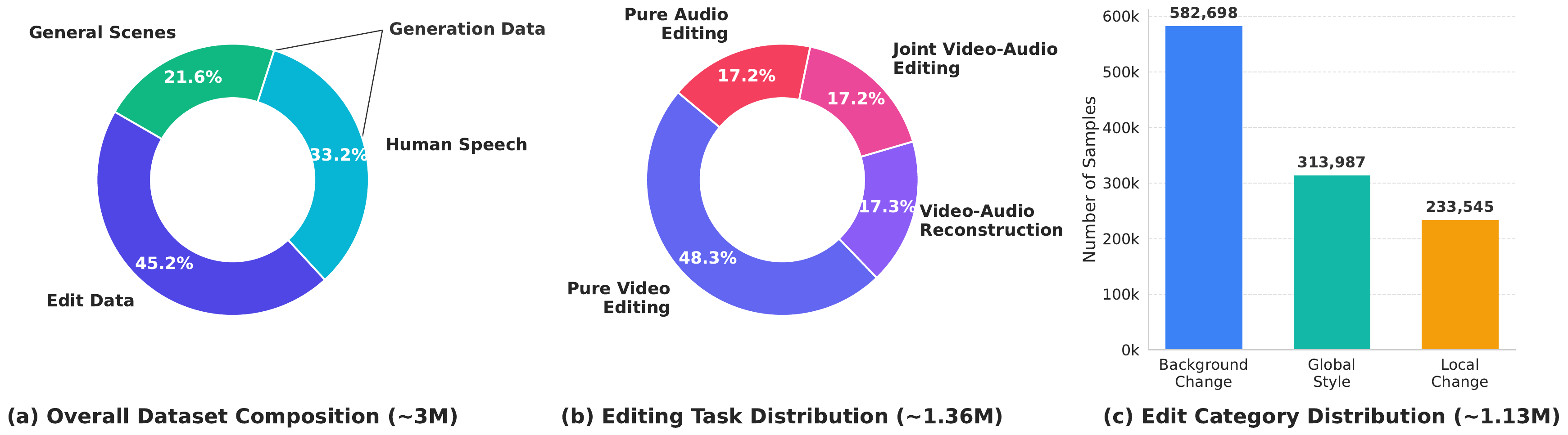}
    \caption{\textbf{Detailed statistics of our video-audio-text dataset.} (a) Overall dataset composition (\(\sim\)3M), highlighting the proportion of Edit Data (45.2\%) versus Generation Data, including General Scenes (21.6\%) and Human Speech (33.2\%). (b) Distribution of the \(\sim\)1.36M editing task samples across four task types. (c) Distribution of the \(\sim\)1.13M edit category samples, excluding reconstruction tasks.}
    \label{fig:dataset_distribution}
\end{figure*}

To train JoVA effectively for both generation and editing tasks, we construct a large-scale, diverse, and highly aligned video-audio-text dataset. As shown in Fig.~\ref{fig:dataset_distribution}, the dataset encompasses approximately 3 million high-quality samples and covers both generation-oriented and editing-oriented data.

\noindent \textbf{Overview of the Training Dataset.}
The overall dataset is broadly categorized into Generation Data and Edit Data, which are further divided into three main subsets:
\begin{itemize}
    \item \textbf{Text-Video-Audio General Scenes (without Speech):} This subset comprises \(\sim\)650K samples (21.6\% of the overall dataset). Sourced from open-source datasets including InternVID~\cite{wang2023internvid}, AudioCaps~\cite{kim2019audiocaps}, and VGGSound~\cite{chen2020vggsound}, it is strictly cleaned and aligned to help the model learn correspondences between visual elements and natural environmental sounds.
    \item \textbf{Text-Video-Audio Human Speech Scenes:} This subset comprises \(\sim\)1M samples (33.2\% of the overall dataset). Collected and annotated by our team, it focuses on strong synchronization among lip movements, facial expressions, and speech in human talking scenarios.
    \item \textbf{Video-Audio Edit Data:} This subset comprises \(\sim\)1.36M samples (45.2\% of the overall dataset). Derived from the human speech scenes, it is constructed using our automated pipeline to generate rich editing instructions and source-target pairs, establishing the foundation for JoVA's editing capabilities.
\end{itemize}
\section{Experiments}

\subsection{Datasets and Evaluation Metrics}

\noindent \textbf{Datasets.} We evaluate our method on three benchmarks. First, we introduce {JoVABench-Gen} and {JoVABench-Edit}, two curated evaluation sets specifically designed to assess video-audio generation and editing capabilities, respectively. We emphasize that all 200 samples in these benchmarks undergo a strict filtering process to ensure there is absolutely no overlap with our training data. Second, to evaluate the broader generalization capabilities of our model, we employ {Verse-Bench}~\cite{UniVerse} as an out-of-domain dataset. It contains 600 image-text prompt pairs spanning a wide range of audio categories, including human speech, animal vocalizations, instrumental music, and natural ambient sounds.

\noindent \textbf{Evaluation Metrics.} Following the evaluation protocol of UniVerse-1~\cite{UniVerse}, we assess our model across multiple dimensions: joint video-audio quality, text-to-speech accuracy, audio generation quality, and video generation fidelity. For video-audio alignment, we measure lip-sync accuracy using SyncNet~\cite{syncnet} to calculate confidence scores (LSE-C) and distances (LSE-D) only on human-speech samples; general-scene videos are strictly excluded because these metrics target visible mouth motion. For general-scene audio-video synchronization, we additionally report DeSync, where lower values indicate better temporal alignment. Text-to-speech quality is evaluated through Word Error Rate (WER) by transcribing generated audio with Whisper-large-v3~\cite{whisper}. Video generation fidelity is comprehensively assessed using Inception Score (IS)~\cite{salimans2016improved}, computed on the generated video frames rather than the audio, for visual quality and diversity, Motion Score (MS) calculated from RAFT-detected~\cite{teed2020raft} optical flow, Aesthetic Score (AS) averaging MANIQA~\cite{yang2022maniqa} fidelity, aesthetic-predictor-v2-5 quality, and Musiq~\cite{ke2021musiq} scores, ID consistency measuring DINOV3~\cite{simeoni2025dinov3} feature similarity between reference images and generated frames, and frame consistency (FC) measuring CLIP similarity between consecutive frames. Additionally, we evaluate text-video semantic consistency using {ImageBind~\cite{girdhar2023imagebind} Text-Video Alignment (IB-TV)}. For audio generation, we evaluate distributional similarity using Fréchet Distance (FD) and Kullback-Leibler (KL) divergence on PANNs~\cite{kong2020panns} and PaSST~\cite{passt} features, semantic consistency via LAION-CLAP~\cite{clap} scores, and AudioBox-Aesthetics~\cite{audiobox} measures for Production Quality (PQ), Production Complexity (PC), Content Enjoyment (CE), and Content Usefulness (CU).

\subsection{Implementation Details}

Unless stated, our framework builds upon the Waver~\cite{zhang2025waver} 12B model as the base architecture.
We adopt a two-stage training strategy: in the first stage, we train the audio branch in isolation on collected text-to-audio and text-to-speech data using a batch size of 256 and a fixed learning rate of $1\times10^{-4}$ for 80K steps. In the second stage, we perform joint video-audio generation and editing training with a batch size of 32 and a learning rate of $3\times10^{-5}$ for 100K steps to learn multimodal interaction. During inference, we apply classifier-free guidance~\cite{cfg} with a weight of 10.0 for both video and audio generation and editing to enhance generation quality.

\subsection{Comparison with State-of-the-Art Models}

On joint video-audio generation benchmarks (JoVABench-Gen and Verse-Bench), we compare our method against two categories of approaches: (1) audio-driven generation methods, where we use ground-truth audio to drive video synthesis for fair comparison of conditional generation capability, including FantasyTalking~\cite{wang2025fantasytalking} and Wan-S2V~\cite{Wan-s2v}; (2) joint video-audio generation methods, including the recent UniVerse-1~\cite{UniVerse}, JavisDiT~\cite{javisdit}, Ovi~\cite{Ovi}, LTX-2~\cite{ltx2}, and UniAVGen~\cite{zhang2025uniavgen} where available.

\begin{table}[t]
\caption{\textbf{Performance comparison on JoVABench-Gen.} The results of other methods are reproduced using their official code. The audio inputs for audio-driven models are from ground-truth audio. Missing reported metrics are denoted by ``-''. $\uparrow$ / $\downarrow$ indicate higher / lower is better.}
\centering
\scriptsize
\renewcommand{\arraystretch}{1.3}
\setlength{\tabcolsep}{2pt}
\resizebox{\linewidth}{!}{%
\begin{tabular}{l *{12}{c}}
\toprule
\textbf{Method}
    & \multicolumn{1}{c}{\textbf{Audio-Video}}
    & \multicolumn{1}{c}{\textbf{TTS}}
    & \multicolumn{7}{c}{\textbf{Audio}}
    & \multicolumn{3}{c}{\textbf{Video}} \\
\cmidrule(lr){2-2}\cmidrule(lr){3-3}\cmidrule(lr){4-10}\cmidrule(lr){11-13}
    & LSE-C$\uparrow$
    & WER$\downarrow$
    & FD$\downarrow$ & KL$\downarrow$ & CS$\uparrow$ & CE$\uparrow$ & CU$\uparrow$ & PC$\downarrow$ & PQ$\uparrow$
    & MS$\uparrow$ & AS$\uparrow$ & ID$\uparrow$ \\
\midrule
\multicolumn{13}{l}{\textit{Audio-driven Generation}} \\
FantasyTalking~\cite{wang2025fantasytalking} &  3.10 & - & - & - & - & - & - & - & - & 0.22 & 0.44  & \textbf{0.87} \\
Wan-S2V~\cite{Wan-s2v}    & 6.43  & - & - & - & - & - & - & - & - & 0.82 & 0.44 & 0.72\\
\midrule
\multicolumn{13}{l}{\textit{Joint Video-Audio Generation}} \\
UniVerse-1~\cite{UniVerse} & 1.62 & 0.37 & 1.04 & 0.83 & 0.16 & 3.68 & 3.90 & 2.12 & 4.39 & 0.43 & 0.42 & 0.82 \\
JavisDiT~\cite{javisdit} & 1.04 & 1.08 & 1.15 & 0.64 & \textbf{0.41} & 3.36 & 3.53 & 2.31 & 4.76 & 0.20 & 0.44 & 0.30 \\
Ovi~\cite{Ovi} & 6.41 & 0.23 & 0.75 & 0.66 & 0.30 & 5.00 & 5.67 & 1.75 & 5.77 & 0.94 & 0.41 & 0.75 \\
LTX-2~\cite{ltx2} & 6.21 & \textbf{0.15} & 0.80 & 0.64 & 0.32 & 5.06 & 5.53 & 1.81 & 5.64 & 0.92 & 0.44 & 0.74 \\
UniAVGen~\cite{zhang2025uniavgen} & 5.87 & 0.17 & 0.70 & \textbf{0.62} & 0.34 & 5.32 & \textbf{6.10} & \textbf{1.62} & 6.30 & 0.90 & 0.44 & 0.75 \\
\rowcolor{cyan!20} \textbf{JoVA} & \textbf{6.70} & 0.19 & \textbf{0.67} & 0.64 & 0.33 & \textbf{5.40} & 6.03 & 1.69 & \textbf{6.49} & \textbf{0.97} & \textbf{0.48} & 0.78 \\
\bottomrule
\end{tabular}%
}
\label{tab:main-jovabench-gen}
\end{table}

\noindent \textbf{JoVABench-Gen.} Table~\ref{tab:main-jovabench-gen} shows that JoVA achieves the best overall performance on JoVABench-Gen. It reaches the highest LSE-C (6.70), surpassing Ovi (6.41), LTX-2 (6.21), UniAVGen (5.87), UniVerse-1 (1.62), and the audio-driven Wan-S2V (6.43), which validates the effect of mouth-area supervision. JoVA also obtains strong speech, audio, and video quality, including the best FD (0.67), CE (5.40), PQ (6.49), MS (0.97), and AS (0.48). On the shared metrics from LTX-2 and UniAVGen, JoVA leads on LSE-C, FD, CE, PQ, MS, AS, and ID while remaining competitive on WER, KL, CU, and PC.

\begin{table*}[ht]
\caption{\textbf{Performance comparison on JoVABench-Edit.} As there are no directly comparable open-source baselines, we combine video editing models with lip-sync models (Diff2Lip~\cite{mukhopadhyay2024diff2lip} and Wav2Lip~\cite{wav2lip}) for comparison, which use ground-truth audio as inputs.}
\centering
\scriptsize
\renewcommand{\arraystretch}{0.98}
\setlength{\tabcolsep}{2pt}
\begin{tabular}{l *{7}{c}}
\toprule
\textbf{Method}
%     & \multicolumn{2}{c}{\textbf{Lip-Sync}}
%     & \multicolumn{5}{c}{\textbf{Video Quality}} \\
% \cmidrule(lr){2-3} \cmidrule(lr){4-8}
    & LSE-C$\uparrow$ & LSE-D$\downarrow$
    & FC$\uparrow$ & IB-TV$\uparrow$ & IS$\uparrow$ & AS$\uparrow$ & MS$\uparrow$ \\
\midrule
\multicolumn{8}{l}{\textit{Combined with Diff2Lip}} \\
Ditto~\cite{ditto} + Diff2Lip~\cite{mukhopadhyay2024diff2lip}      & 2.44 & 5.30 & 0.98 & \textbf{0.23} & 2.13 & 0.34 & 0.35 \\
ICVE~\cite{icve} + Diff2Lip~\cite{mukhopadhyay2024diff2lip}       & 2.45 & 5.35 & \textbf{0.99} & 0.18 & 1.72 & 0.36 & 0.30 \\
InsViE~\cite{wu2025insvie} + Diff2Lip~\cite{mukhopadhyay2024diff2lip}     & 2.38 & 5.72 & \textbf{0.99} & 0.21 & 1.02 & 0.39 & 0.28 \\
Lucy~\cite{lucyedit} + Diff2Lip~\cite{mukhopadhyay2024diff2lip}       & 2.34 & 5.83 & 0.98 & 0.18 & 3.01 & 0.28 & 0.30 \\
OmniVideo~\cite{omnivideo} + Diff2Lip~\cite{mukhopadhyay2024diff2lip}  & 4.07 & 5.50 & 0.98 & 0.19 & 3.03 & 0.35 & \textbf{0.71} \\
\midrule
\multicolumn{8}{l}{\textit{Combined with Wav2Lip}} \\
Ditto~\cite{ditto} + Wav2Lip~\cite{wav2lip}       & 5.85 & 2.83 & 0.98 & \textbf{0.23} & 2.33 & 0.42 & 0.46 \\
ICVE~\cite{icve} + Wav2Lip~\cite{wav2lip}        & 5.54 & 2.50 & 0.98 & 0.15 & 1.55 & 0.41 & 0.32 \\
InsViE~\cite{wu2025insvie} + Wav2Lip~\cite{wav2lip}      & 5.16 & 4.40 & 0.97 & 0.18 & 1.36 & 0.37 & 0.24 \\
Lucy~\cite{lucyedit} + Wav2Lip~\cite{wav2lip}        & 5.37 & 3.24 & 0.98 & 0.19 & 3.01 & 0.34 & 0.53 \\
OmniVideo~\cite{omnivideo} + Wav2Lip~\cite{wav2lip}   & 4.86 & 1.75 & 0.97 & 0.18 & 2.27 & 0.37 & 0.63 \\
\midrule
\rowcolor{cyan!20}\textbf{JoVA} & {\textbf{5.88}} & {\textbf{1.66}} & {0.98} & {0.21} & {\textbf{3.21}} & {\textbf{0.47}} & {0.66} \\
\bottomrule
\end{tabular}%
\label{tab:lip-sync-comparison}
\end{table*}

\noindent \textbf{JoVABench-Edit.} Table~\ref{tab:lip-sync-comparison} evaluates JoVA on JoVABench-Edit. Since no directly comparable open-source baseline exists for joint video-audio editing, we construct cascaded pipelines that combine video editing models with Diff2Lip or Wav2Lip using ground-truth audio. These baselines therefore solve a strictly easier conditional task, while JoVA jointly synthesizes both video and audio. Despite this advantage, JoVA achieves the best lip-sync accuracy (LSE-C \(5.88\), LSE-D \(1.66\)) and the strongest video fidelity (IS \(3.21\), AS \(0.47\)).

\begin{table*}[!t]
\caption{\textbf{Performance comparison on Verse-Bench.} The results of Ovi and JavisDiT are reproduced using their official code. Missing reported metrics are denoted by ``-''. $\uparrow$ / $\downarrow$ indicate higher / lower is better.}
\centering
\scriptsize
\renewcommand{\arraystretch}{1.05}
\setlength{\tabcolsep}{3pt}
\resizebox{\textwidth}{!}{%
\begin{tabular}{l *{13}{c}}
\toprule
\textbf{Method}
    & \multicolumn{2}{c}{\textbf{Audio-Video}}
    & \textbf{TTS}
    & \multicolumn{7}{c}{\textbf{Audio}}
    & \multicolumn{3}{c}{\textbf{Video}} \\
\cmidrule(lr){2-3} \cmidrule(lr){4-4} \cmidrule(lr){5-11} \cmidrule(lr){12-14}
    & LSE-C$\uparrow$
    & DeSync$\downarrow$
    & WER$\downarrow$
    & FD$\downarrow$ & KL$\downarrow$ & CS$\uparrow$ & CE$\uparrow$ & CU$\uparrow$ & PC$\downarrow$ & PQ$\uparrow$
    & MS$\uparrow$ & AS$\uparrow$ & ID$\uparrow$ \\
\midrule
\multicolumn{14}{l}{\textit{Audio-driven Generation}} \\
FantasyTalking~\cite{wang2025fantasytalking} & 2.68 & - & - & - & - & - & - & - & - & - & 0.07 & 0.42 & 0.87 \\
Wan-S2V~\cite{Wan-s2v}        & 6.49 & - & - & - & - & - & - & - & - & - & 0.17 & 0.46 & 0.89 \\
\midrule
\multicolumn{14}{l}{\textit{Joint Video-Audio Generation}} \\
UniVerse-1~\cite{UniVerse}      & 1.62 & 0.23 & 0.18 & 1.25 & 2.70 & 0.16 & 3.53 & 4.61 & \textbf{2.49} & 5.20 & 0.20 & 0.47 & \textbf{0.89} \\
JavisDiT~\cite{javisdit} & 0.85 & 0.27 & 1.00 & 0.99 & 1.75 & 0.19 & 3.48 & 5.03 & 2.49 & 5.49 & 0.27 & 0.42 & 0.40 \\
Ovi~\cite{Ovi}           & {6.61} & 0.49 & 0.12 & 0.97 & 2.23 & 0.20 & 3.95 & {5.60} & 2.09 & {6.20} & 0.55 & 0.47 & 0.86 \\
\rowcolor{cyan!20}\textbf{JoVA} & \textbf{6.65} & \textbf{0.16} & \textbf{0.11} & \textbf{0.82} & \textbf{1.39} & \textbf{0.29} & \textbf{4.47} & \textbf{5.94} & 2.47 & \textbf{6.23}& \textbf{0.75} & \textbf{0.49} & 0.86 \\
\bottomrule
\end{tabular}%
}
\label{tab:main-results}
\end{table*}

JoVA does not rank first in every single metric, but it provides the best overall trade-off. Its MS of \(0.66\) is close to OmniVideo+Diff2Lip (\(0.71\)), while avoiding that pipeline's lower lip-sync score (LSE-C \(4.07\)). It also maintains competitive IB-TV (\(0.21\)) and FC (\(0.98\)) without sacrificing video quality (IS/AS), showing the advantage of avoiding error accumulation in two-stage pipelines.

\noindent \textbf{Verse-Bench.} On Verse-Bench (Table~\ref{tab:main-results}), JoVA remains strong across diverse scenarios. It achieves the lowest WER (0.11), competitive LSE-C (6.65), the best DeSync (0.16), FD (0.82), KL (1.39), CS (0.29), CE (4.47), MS (0.75), and AS (0.49), while preserving identity consistency (0.86).

\subsection{Diagnostic Study}

\begin{table*}[t]
\centering
\scriptsize
\renewcommand{\arraystretch}{1.08}
\setlength{\tabcolsep}{1pt}

\begin{minipage}{0.45\textwidth}
\centering
\caption{\textbf{Effect of mouth-area loss weight \(\lambda\) on JoVABench-Gen.} $\uparrow$ / $\downarrow$ indicate higher / lower is better.}
\begin{tabular}{ccccccc}
\toprule
\(\lambda\)  & LSE-C\(\uparrow\) & WER\(\downarrow\) & FD\(\downarrow\) & KL\(\downarrow\) & PQ\(\uparrow\) & AS\(\uparrow\) \\
\midrule
0.0 & 1.39 & 0.17 & 0.67 & 0.63 & 6.44 & 0.48 \\
2.0 & 6.53 & 0.18 & 0.67 & 0.62 & 6.47 & 0.46 \\
\rowcolor{cyan!20}5.0 & 6.70 & 0.19 & 0.67 & 0.64 & 6.49 & 0.48 \\
8.0 & 6.65 & 0.18 & 0.68 & 0.63 & 6.40 & 0.47 \\
\bottomrule
\end{tabular}
\label{tab:ablation_mouth}
\end{minipage}%
\hfill
\begin{minipage}{0.52\textwidth}
\centering
\caption{\textbf{Comparison of audio-video fusion strategies on JoVABench-Gen.} All variants share the same backbone capacity, training data, and schedule; only the cross-modal interaction mechanism differs. $\uparrow$ / $\downarrow$ indicate higher / lower is better.}
\resizebox{\linewidth}{!}{%
\begin{tabular}{lccccc}
\toprule
\textbf{Strategy} & LSE-C\(\uparrow\) & WER\(\downarrow\) & PQ\(\uparrow\) & AS\(\uparrow\) & ID\(\uparrow\) \\
\midrule
Seq. Cross-Attn. (Ovi-style) & 6.51 & 0.24 & 6.25 & 0.48 & 0.75 \\
Parallel Cross-Attn. (w/ linear) & 1.59 & 0.29 & 6.31 & 0.46 & 0.74 \\
Parallel Cross-Attn. (w/o linear) & 6.57 & 0.28 & 6.14 & 0.47 & 0.76 \\
\rowcolor{cyan!20} Joint Self-Attn. (Ours) & \textbf{6.70} & \textbf{0.19} & \textbf{6.49} & \textbf{0.48} & \textbf{0.78} \\
\bottomrule
\end{tabular}
}
\label{tab:attention_comparison}
\end{minipage}
\end{table*}

\noindent \textbf{Effectiveness of Mouth-Area Loss.}
We first investigate the impact of the mouth-area loss weight \(\lambda\) in Eq.~\eqref{eq:total_loss}. As shown in Table~\ref{tab:ablation_mouth}, increasing \(\lambda\) from 0.0 dramatically improves lip-sync accuracy (LSE-C from 1.39 to 6.53), demonstrating the critical role of targeted supervision on the mouth region. Without mouth-area loss (\(\lambda=0.0\)), the model does not fully learn fine-grained lip-speech alignment through joint attention alone. The optimal performance is achieved at \(\lambda=5.0\) (LSE-C: 6.70), where the model effectively balances lip-sync precision with overall generation quality. Further increasing \(\lambda\) to 8.0 leads to marginal degradation (LSE-C: 6.65). Notably, audio quality metrics (WER, KL, PQ) and AS remain stable across different \(\lambda\) values, confirming that our mouth-area supervision strategy enhances lip-sync without sacrificing general audio-visual quality. The substantial improvement from \(\lambda=0.0\) to \(\lambda=2.0\) (LSE-C: 1.39 \(\rightarrow\) 6.53) highlights the necessity of explicit mouth-region guidance for achieving precise synchronization.

\begin{figure*}[t]
    \centering
    \includegraphics[width=0.98\linewidth]{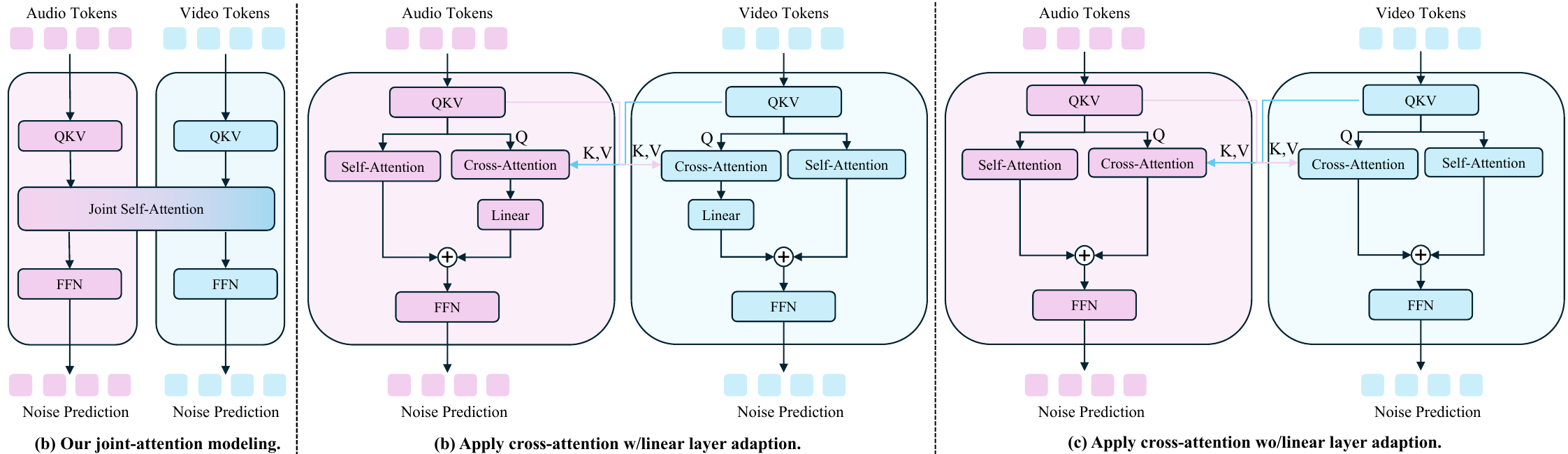}
    \caption{\textbf{Comparison of different audio-video fusion strategies.} (a) Joint self-attention processes video, audio, and text tokens together in one layer. (b,c) Cross-attention variants exchange information through explicit audio-video attention paths, while text conditioning remains inside the self-attention layers.}
    \label{fig:supp_cross_attention}
\end{figure*}

\noindent \textbf{Comparison of Joint Self-Attention and Cross-Attention}.
To validate our joint self-attention design, we compare it with three cross-attention variants under identical capacity and training data. As shown in Fig.~\ref{fig:supp_cross_attention}, JoVA processes video, audio, and text tokens together in unified self-attention, enabling bidirectional multimodal interaction while reusing pretrained transformer parameters. In contrast, the sequential Ovi-style variant alternates cross-modal interaction, while the two parallel variants compute cross-modal attention alongside self-attention and merge the results either with or without an additional linear adaptation layer.

Tab.~\ref{tab:attention_comparison} shows that joint self-attention achieves the best LSE-C (6.70), WER (0.19), PQ (6.49), ID (0.78), and matching-best AS (0.48). The low LSE-C of the parallel cross-attention variant with linear adaptation (1.59) suggests that the added projection can disrupt cross-modal alignment. These results validate that unified multimodal processing is essential for high-quality, synchronized audio-visual generation without introducing additional parameters.

\subsection{Qualitative Results}
\begin{figure}[H]
    \centering
    \includegraphics[width=0.85\linewidth]{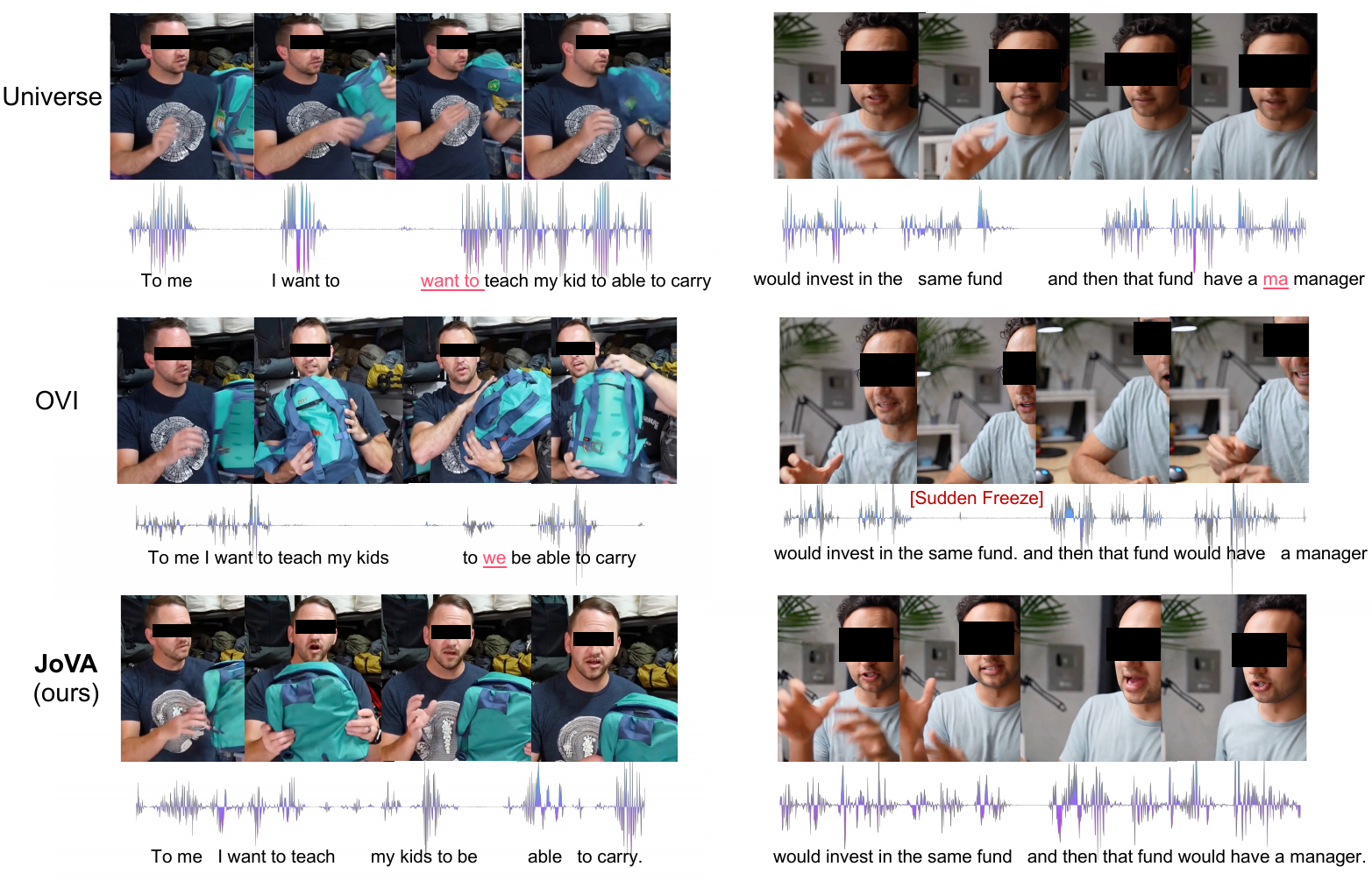}
    \caption{\textbf{Qualitative comparison of speech generation and lip-sync quality.} Generated video frames with audio waveforms and ASR transcriptions. Red underlines indicate speech recognition errors. UniVerse-1 omits words, Ovi shows freezing artifacts and mispronunciations, while JoVA produces accurate speech with precise lip-sync alignment.}
\label{fig:qualitative}
\end{figure}

Fig.~\ref{fig:qualitative} presents qualitative comparisons of video-audio generation quality. We visualize generated frames alongside their audio waveforms and automatic speech recognition (ASR) results, with recognition errors underlined in red. UniVerse-1 produces incomplete speech with missing words, indicating weaker audio generation quality. Ovi shows temporal consistency issues, including sudden freezing artifacts and mispronunciations. In contrast, JoVA generates smooth, natural speech with accurate word pronunciation and maintains precise lip-speech synchronization throughout the sequence. The aligned audio waveforms and correct ASR transcriptions demonstrate JoVA's superior capability in joint video-audio generation with proper cross-modal alignment.

\begin{figure*}[!p]
    \centering
    \includegraphics[width=0.82\textwidth]{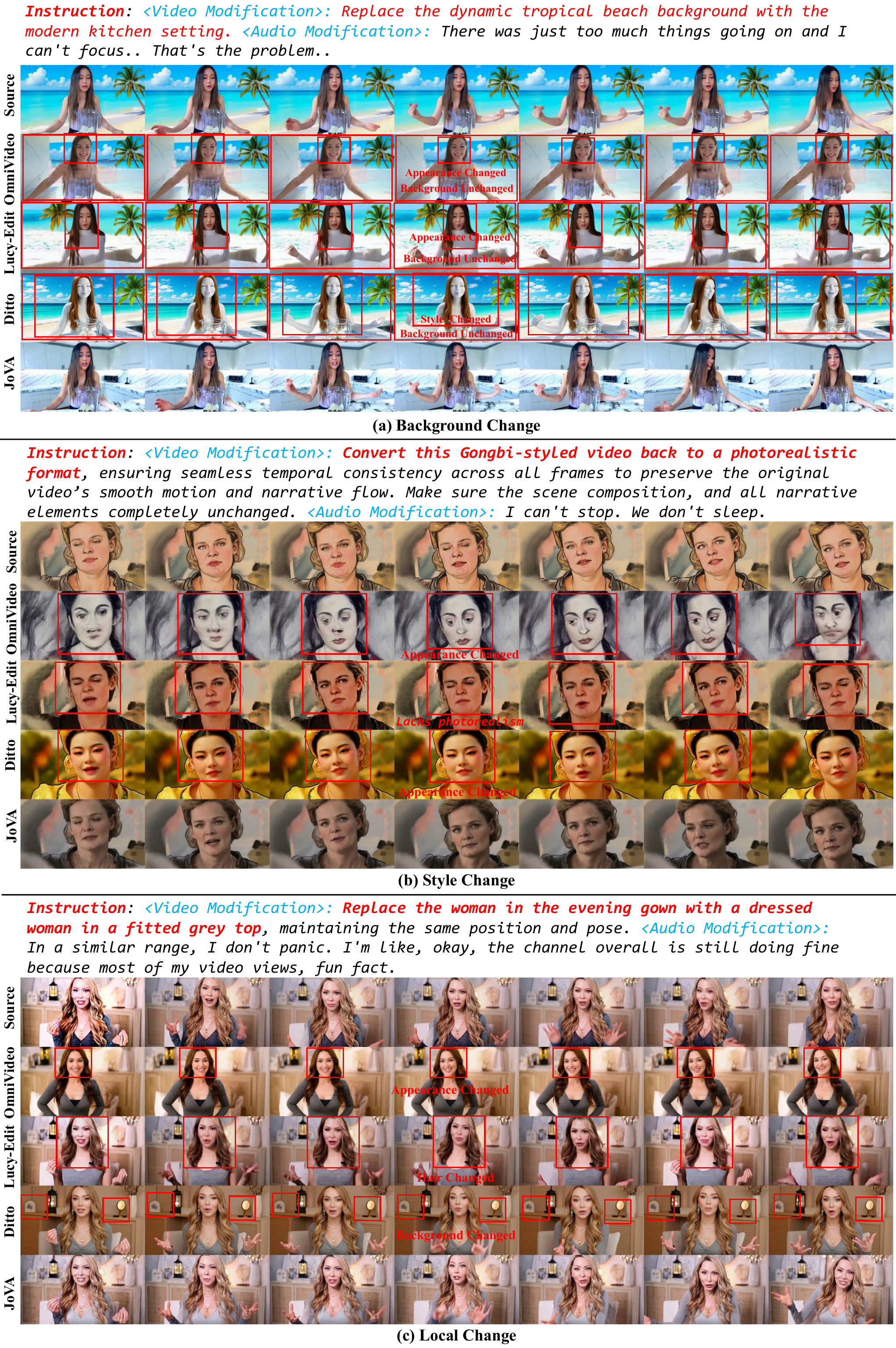}
    \caption{\textbf{Qualitative comparison of video-audio editing.} We compare JoVA against cascaded baselines (OmniVideo + Diff2Lip and Ditto + Wav2Lip, with ground-truth audio) across background change, style change, and local change. JoVA better follows the editing instructions while preserving unedited regions, whereas the baseline pipelines often introduce identity shifts, attribute entanglement, or incomplete edits. Pose differences are expected because the edited subjects are driven by modified audio inputs.}
    \label{fig:edit_comparison}
\end{figure*}

Fig.~\ref{fig:edit_comparison} further compares video-audio editing. JoVA better follows background, style, and local edit instructions while preserving unedited regions, whereas baselines tend to introduce identity shifts, entangle unrelated attributes, or produce incomplete edits under the same audio-conditioned setting. For background replacement, JoVA changes the scene while preserving the subject appearance; for style conversion, it restores photorealism without corrupting temporal consistency; for local replacement, it edits target attributes without changing the global layout. These cases show that joint video-audio conditioning provides stronger instruction following than cascaded editing baselines. More visualizations are given in the supplementary materials.

\section{Conclusion}
In this paper, we introduce JoVA, an extensible framework that unifies joint video-audio generation and editing. Native joint self-attention eliminates complex task-specific fusion modules and enables direct interaction among video, audio, and text, while channel-wise conditioning supports flexible visual editing without prohibitive token growth. A simple mouth-area loss further improves lip-speech synchronization for human-centric content.

To train and evaluate JoVA, we curate a comprehensive multi-domain corpus and benchmarks. Experiments show state-of-the-art performance across generation and editing tasks, especially in multimodal alignment and generation quality. We also observe several limitations: very fast hand motion can still induce finger artifacts in generation, and instruction-based editing does not always preserve text in text-rich backgrounds. In addition, our audio-video editing currently focuses on human-speech scenes, as paired audio-video editing data for general scenes remains difficult to obtain; this is a data limitation rather than an architectural one, and our framework can be extended to general-scene editing once suitable data becomes available. Future work will address these issues by expanding training diversity and broadening JoVA's coverage of high-motion scenes, text-rich backgrounds, and general-scene audio-video editing. JoVA establishes a strong foundation for unified multimodal content creation.

\section{Declaration}
All datasets, images and videos of designed character portraits used in this paper are solely for research purposes. In addition, the datasets and models employed in this paper are also exclusively for research use and will not be integrated into any products of ByteDance.

\section*{Acknowledgments}
This work is supported by Hong Kong Research Grants Council -- General Research Fund (Grant No.\ 17211024), Hong Kong Innovation and Technology Commission -- Innovation and Technology Fund (Grant No.\ ITS/488/24FP), and HKU Seed Fund for PI Research.

\clearpage
\bibliographystyle{plainnat}
\bibliography{main}

\clearpage
\beginappendix
\section{Quantitative Evaluation on Video Generation}
\label{sec:vbench_eval}

In addition to evaluating the joint video-audio capabilities, we also assess the pure video generation performance of our method on the widely recognized VBench~\cite{Vbench} benchmark. This allows us to rigorously compare JoVA's visual generation quality against several state-of-the-art video generation models. 

As shown in Table~\ref{tab:vbench_selected_comparison}, JoVA achieves the highest {Total Score (82.61)} among the compared methods, surpassing strong baselines such as Gen-3~\cite{gen3} (82.32), Kling~\cite{kuaishou2024kling} (81.85), and CogVideoX-5B~\cite{yang2024cogvideox} (81.61). Furthermore, our model demonstrates exceptional capability in complex compositional generation and motion synthesis, securing the top scores in {Semantic Score (77.47)}, {Dynamic Degree (78.33)}, and {Multiple Objects (73.20)}. It also maintains highly competitive performance in Quality Score (83.90) and Human Action (99.20), closely trailing leading commercial and open-source models. On Appearance Style (22.17), JoVA is in the same range as the other open-source baselines but is not the best, reflecting our focus on high-fidelity content generation rather than stylized rendering. Overall, these results indicate that while JoVA is designed for unified video-audio generation and editing, its foundational video generation quality remains at the forefront of the field, effectively handling complex semantics and highly dynamic scenes.

\begin{table*}[htbp]
\centering
\caption{Comparison with state-of-the-art video generation models on VBench~\cite{Vbench}. The results of other methods are borrowed from~\cite{chen2025goku}.}
\label{tab:vbench_selected_comparison}
\resizebox{\textwidth}{!}{%
\begin{tabular}{l ccc ccccc}
\toprule
Method & 
\rotatebox{45}{Total Score} & 
\rotatebox{45}{Quality Score} & 
\rotatebox{45}{Semantic Score} & 
\rotatebox{45}{Human Action} & 
\rotatebox{45}{Scene} & 
\rotatebox{45}{Dynamic Degree} & 
\rotatebox{45}{Multiple Objects} & 
\rotatebox{45}{Appear. Style} \\
\midrule
AnimateDiff-V2~\cite{Animatediff}   & 80.27 & 82.90 & 69.75 & 92.60 & 50.19 & 40.83 & 36.88 & 22.42 \\
VideoCrafter-2.0~\cite{chen2024videocrafter2} & 80.44 & 82.20 & 73.42 & 95.00 & \textbf{55.29} & 42.50 & 40.66 & \textbf{25.13} \\
OpenSora V1.2~\cite{lin2024open}    & 79.23 & 80.71 & 73.30 & 85.80 & 42.47 & 47.22 & 58.41 & 23.89 \\
Show-1~\cite{Show-1}           & 78.93 & 80.42 & 72.98 & 95.60 & 47.03 & 44.44 & 45.47 & 23.06 \\
Gen-3~\cite{gen3}            & 82.32 & \textbf{84.11} & 75.17 & 96.40 & 54.57 & 60.14 & 53.64 & 24.31 \\
Pika-1.0~\cite{pika2024pika}         & 80.69 & 82.92 & 71.77 & 86.20 & 49.83 & 47.50 & 43.08 & 22.26 \\
CogVideoX-5B~\cite{yang2024cogvideox}     & 81.61 & 82.75 & 77.04 & \textbf{99.40} & 53.20 & 70.97 & 62.11 & 24.91 \\
Kling~\cite{kuaishou2024kling}            & 81.85 & 83.39 & 75.68 & 93.40 & 50.86 & 46.94 & 68.05 & 19.62 \\
Mira~\cite{ju2024miradata}            & 71.87 & 78.78 & 44.21 & 63.80 & 16.34 & 60.33 & 12.52 & 21.89 \\
\midrule
\textbf{JoVA (Ours)} & \textbf{82.61} & {83.90} & \textbf{77.47} & {99.20} & {49.36} & \textbf{78.33} & \textbf{73.20} & {22.17} \\
\bottomrule
\end{tabular}%
}
\end{table*}

\section{Human Subjective Evaluation}
\label{sec:supp_human_eval}

To rigorously assess the perceptual quality and alignment of our generated and edited outputs, we conduct a comprehensive human subjective evaluation. The user study involves 21 participants who are tasked with evaluating 50 diverse samples for each task. Participants blindly rate the outputs on a Mean Opinion Score (MOS) scale ranging from 1 (worst) to 5 (best).

For the \textbf{Video-Audio Generation} task (Table~\ref{tab:mos_generation}), users evaluate the models across three key dimensions: Video Quality, Audio Quality, and Audio-Video Synchronization. As shown in the results, JoVA consistently outperforms the baseline methods (UniVerse-1~\cite{UniVerse} and Ovi~\cite{Ovi}) in all metrics, demonstrating superior capability in generating high-fidelity, well-synchronized multimodal content.

\begin{table*}[t]
    \centering
    \renewcommand{\arraystretch}{1.1}

    \begin{minipage}[t]{0.45\textwidth}
        \caption{\textbf{Human Subjective Evaluation on Video-Audio Generation.} Mean Opinion Scores (MOS) on a 1--5 scale.}
        \label{tab:mos_generation}
    \end{minipage}
    \hfill
    \begin{minipage}[t]{0.52\textwidth}
        \caption{\textbf{Human Subjective Evaluation on Video-Audio Editing.} MOS on a 1--5 scale. Audio Quality is omitted as the baseline methods utilize ground-truth audio as input.}
        \label{tab:mos_editing}
    \end{minipage}
    
    \vspace{2mm}

    \begin{minipage}[t]{0.45\textwidth}
        \centering
        \resizebox{\linewidth}{!}{%
        \begin{tabular}{lccc}
        \toprule
        Model & Video Quality & Audio Quality & A-V Sync \\
        \midrule
        UniVerse-1~\cite{UniVerse} & 3.65 & 2.85 & 2.10 \\
        Ovi~\cite{Ovi} & 3.95 & 3.65 & 4.05 \\
        \textbf{JoVA (Ours)} & \textbf{4.05} & \textbf{4.15} & \textbf{4.10} \\
        \bottomrule
        \end{tabular}%
        }
    \end{minipage}
    \hfill
    \begin{minipage}[t]{0.52\textwidth}
        \centering
        \resizebox{\linewidth}{!}{%
        \begin{tabular}{lccc}
        \toprule
        Model & Video Quality & A-V Sync & Instruct-Follow \\
        \midrule
        OmniVideo~\cite{omnivideo} + Diff2Lip~\cite{mukhopadhyay2024diff2lip} & 2.85 & 2.95 & 2.70 \\
        Ditto~\cite{ditto} + Wav2Lip~\cite{wav2lip}      & 3.40 & 3.75 & 3.30 \\
        \textbf{JoVA (Ours)} & \textbf{3.95} & \textbf{4.00} & \textbf{3.85} \\
        \bottomrule
        \end{tabular}%
        }
    \end{minipage}
\end{table*}

For the \textbf{Video-Audio Editing} task (Table~\ref{tab:mos_editing}), we introduce an additional metric, \textit{Instruction-Following}, to measure how accurately the models execute the given text prompts while preserving the unedited regions. Note that we omit the Audio Quality metric for this task, as the baseline methods directly utilize ground-truth audio as input. We compare JoVA against OmniVideo~\cite{omnivideo} + Diff2Lip~\cite{mukhopadhyay2024diff2lip} and a much stronger baseline pipeline, Ditto~\cite{ditto} + Wav2Lip~\cite{wav2lip}. While Ditto~\cite{ditto} + Wav2Lip~\cite{wav2lip} yields competitive visual quality, JoVA still maintains a clear advantage. JoVA achieves a higher score in Video Quality (\textbf{3.95} vs. 3.40), A-V Sync (\textbf{4.00} vs. 3.75), and Instruction-Following (\textbf{3.85} vs. 3.30), highlighting its robustness in handling complex, multimodal editing instructions without compromising temporal alignment.

\section{Additional Dataset Details}
\label{sec:dataset_details}

\subsection{Details of Video-Audio Generation Data}

\begin{figure}[t]
    \centering
    \includegraphics[width=\linewidth]{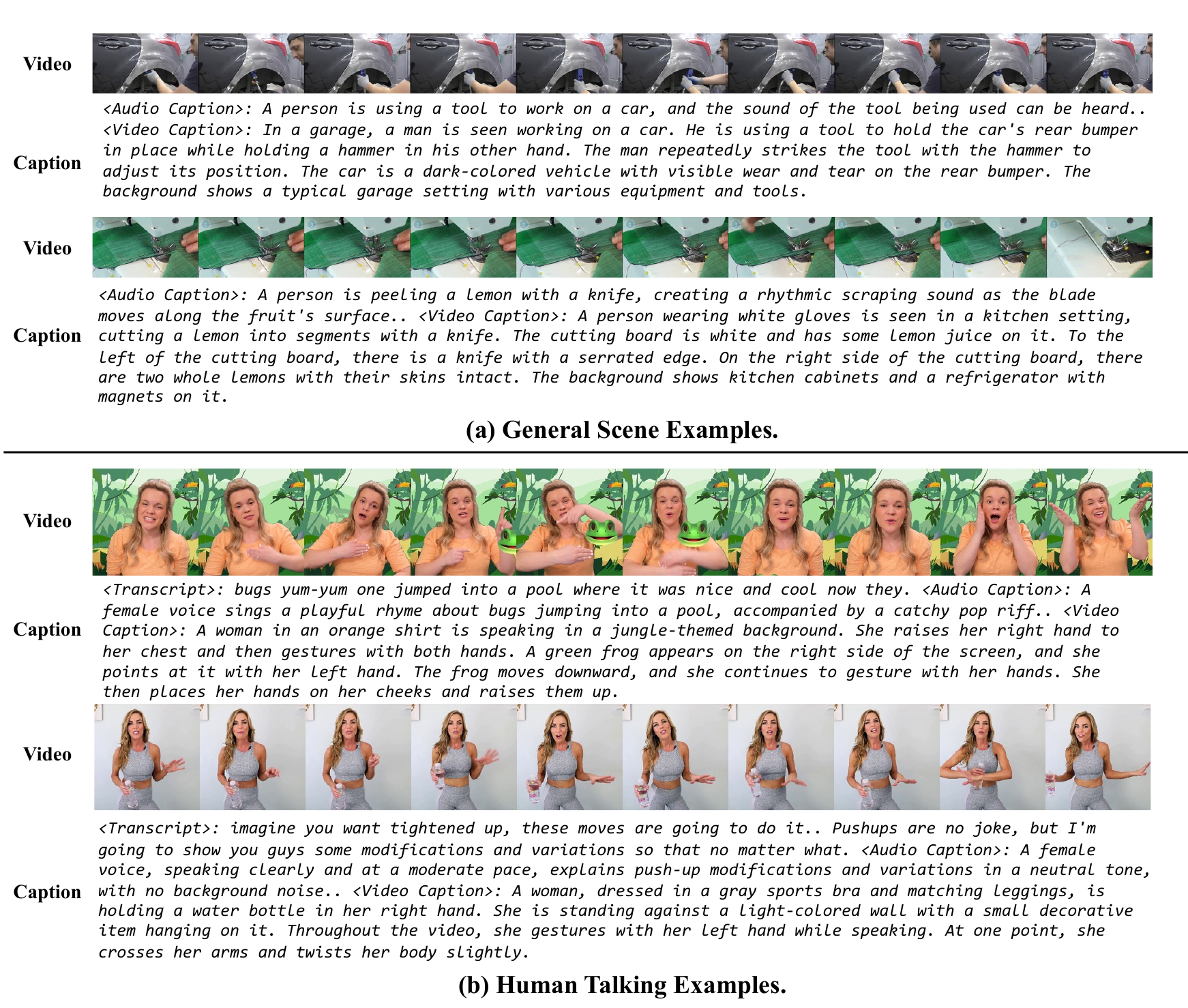}
    \caption{Examples of the video-audio generation training data. The samples showcase diverse scenarios, including general open-domain scenes and human speech environments, accompanied by their corresponding text descriptions used for generation tasks.}
    \label{fig:generation_examples}
\end{figure}

The generation subsets (General Scenes and Human Speech Scenes) serve as the foundation for the model's ability to synthesize high-fidelity, synchronized video and audio from textual prompts. To provide an intuitive understanding of this data, Figure \ref{fig:generation_examples} illustrates typical training examples used during the generation phase. These examples highlight the diversity of our dataset, capturing both complex open-domain environmental dynamics and fine-grained human facial movements, ensuring the model learns robust modality-specific representations before advancing to complex editing tasks.

\subsection{Detailed Distribution of Video-Audio Edit Data}

To ensure robust and precise video-audio editing capabilities, our Edit Data encompasses a wide variety of tasks and granular categories.

\begin{figure}[h!]
    \centering
    \includegraphics[width=\linewidth]{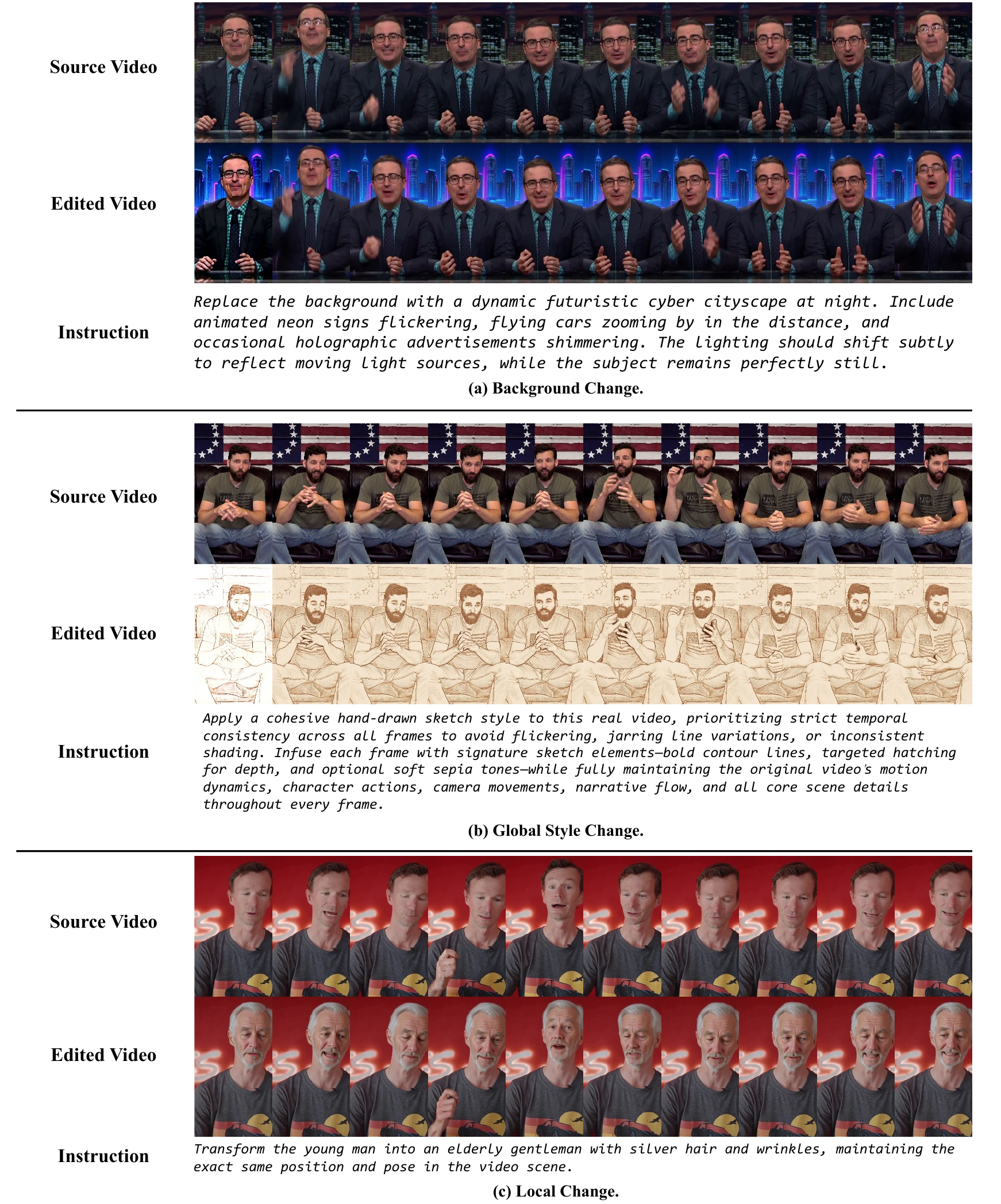}
    \caption{Examples of the three fine-grained edit categories present in our training dataset: Background Change, Global Style, and Local Change.}
    \label{fig:edit_categories}
\end{figure}

\noindent\textbf{Task Distribution:} 
The edit dataset contains a total of 1,367,438 samples distributed across four distinct task types: pure video editing, pure audio editing, joint video-audio editing, and video-audio reconstruction. Pure video editing constitutes the largest portion at approximately 48.3\%, while the remaining three tasks are evenly distributed, each accounting for around 17\% of the total data. This balanced composition allows the model to master both independent modality manipulation and joint multimodal editing.

\noindent\textbf{Edit Category Distribution:} 
For the actual editing tasks (excluding the reconstruction data, leaving a total of 1,130,230 samples), we further categorize the editing instructions into three fine-grained levels: Background Change, Global Style, and Local Change. As visually demonstrated in Figure \ref{fig:edit_categories}, this hierarchical categorization equips the model to handle diverse editing demands, ranging from macroscopic stylistic transformations to microscopic local modifications.

Through the aforementioned rich and well-distributed dataset, \textbf{JoVA} is able to fully learn the alignment relationships between video and audio modalities, demonstrating excellent generalization capabilities in complex joint generation and editing tasks.

\begin{table}[t]
    \centering
    \caption{\textbf{Training configurations for the two-stage training strategy of JoVA.} Stage 1 trains the audio branch in isolation; Stage 2 performs joint video-audio generation and editing.}
    \label{tab:training_configs}
    \resizebox{\linewidth}{!}{%
    \begin{tabular}{l c c c c}
        \toprule
        \textbf{Stage} & \textbf{Training Data / Task} & \textbf{Batch Size} & \textbf{Learning Rate} & \textbf{Training Steps} \\
        \midrule
        Stage 1 & Text-to-Audio \& Text-to-Speech & 256 & \(1\times10^{-4}\) & 80K \\
        Stage 2 & Joint Generation \& Editing & 32 & \(3\times10^{-5}\) & 100K \\
        \bottomrule
    \end{tabular}%
    }
\end{table}

\section{Training Strategy and Configurations}
\label{sec:training_strategy}

To effectively learn both modality-specific representations and complex multimodal interactions, we adopt a two-stage training strategy for \textbf{JoVA}.

\noindent\textbf{Stage 1: Audio Branch Initialization and Modality Training.}
In the first stage, we initialize the audio branch by duplicating the structural design of the pretrained video model. We then train this individual audio modality branch exclusively on the text-to-audio and text-to-speech samples from the aforementioned video-audio Generation Data. This allows the audio branch to develop strong generative capabilities before being integrated with the visual modality.

\noindent\textbf{Stage 2: Joint Multi-modal Training.}
In the second stage, we perform joint video-audio generation and editing training. By utilizing both the video-audio generation data and the video-audio editing data, the model learns complex multimodal interactions, enabling it to perform synchronized generation and precise editing tasks.

The detailed hyperparameters for both training stages are summarized in Table \ref{tab:training_configs}.

\section{Mouth-Area Mask Validation}
\label{sec:mouth-area}

\begin{figure}[h]
    \centering
    \includegraphics[width=0.8\linewidth]{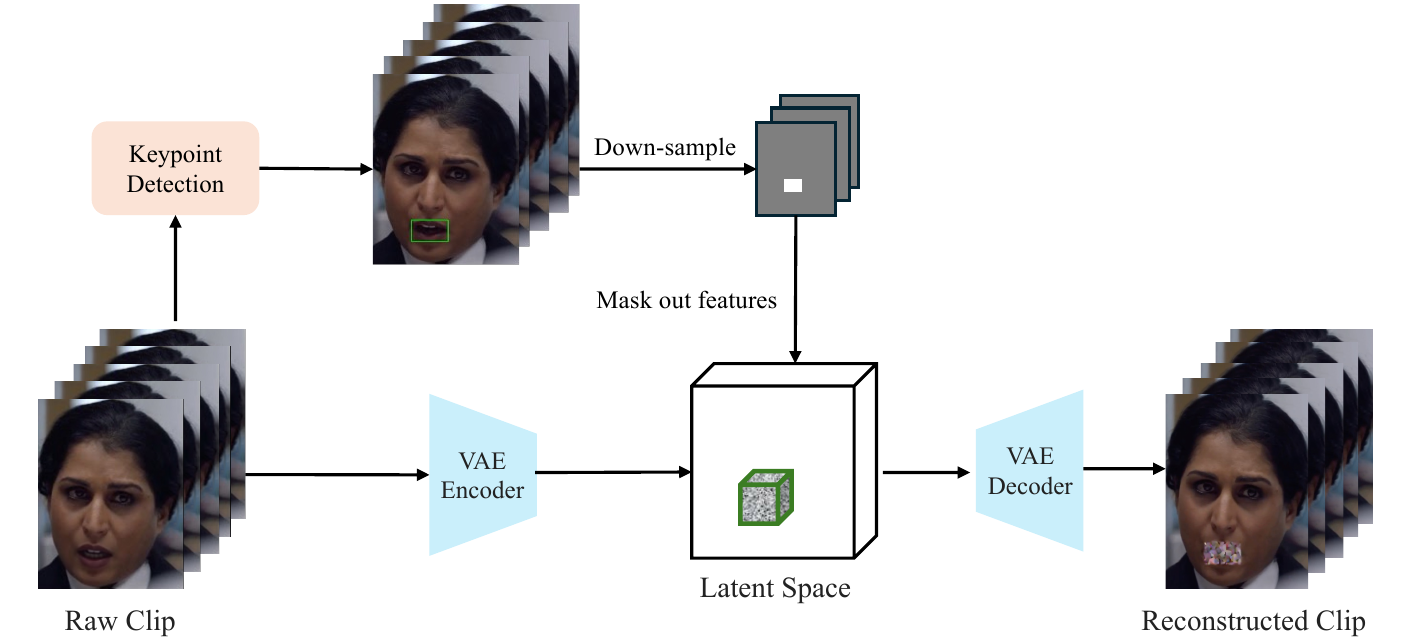}
    \caption{\textbf{Mouth-area mask generation and validation pipeline.} Detected mouth regions are mapped to VAE latent space, then zero-masked and decoded to verify localization accuracy through selective visual degradation.}
    \label{fig:lip_mask}
\end{figure}

\begin{figure}[t!]
    \centering
    \includegraphics[width=0.9\linewidth]{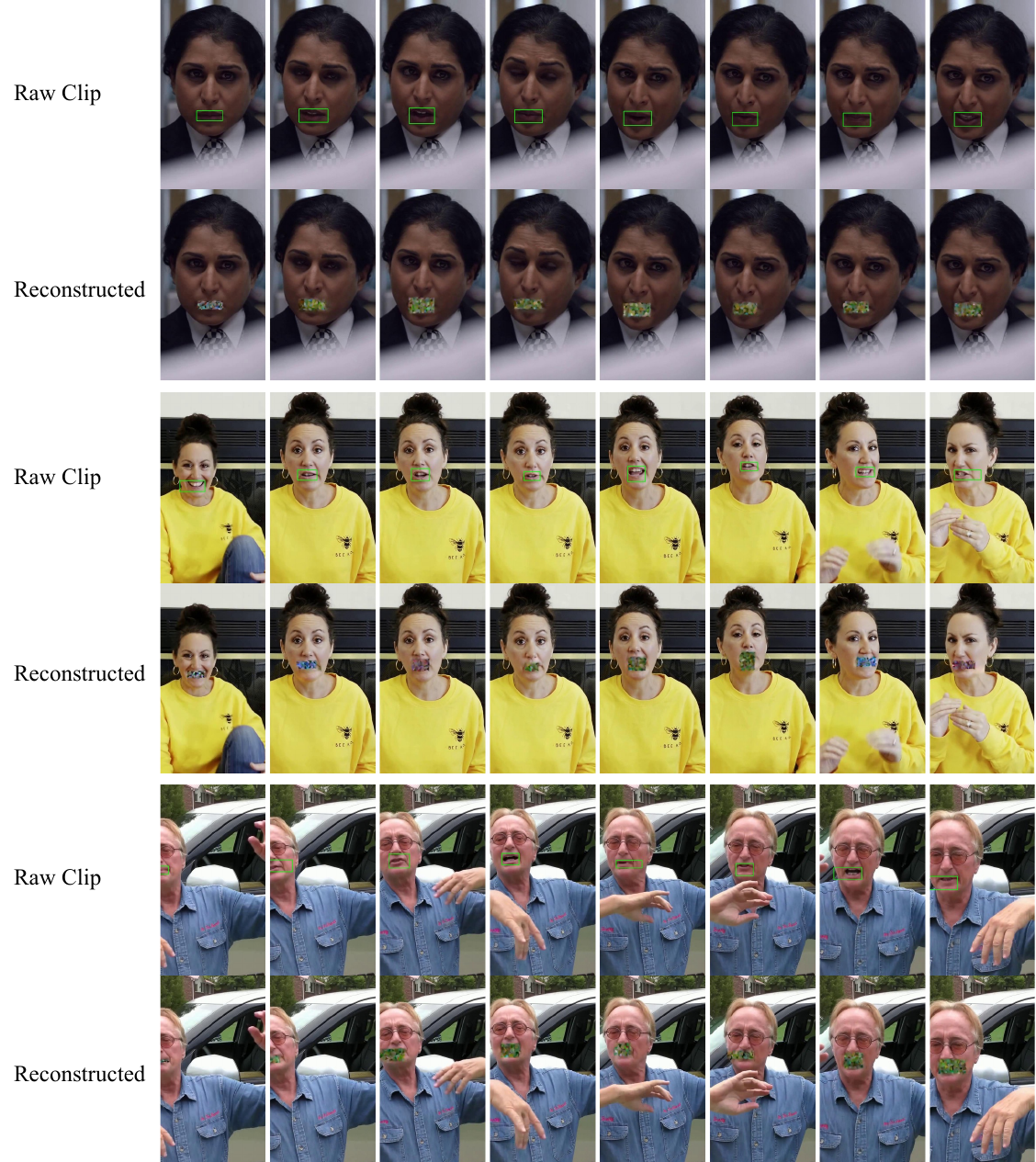}
    \caption{\textbf{Zero-masking validation results across four video samples.} For each sample, the top row shows the original video frames with detected mouth bounding boxes (green), and the bottom row shows the reconstructed frames after zero-masking the corresponding region in VAE latent space. The selective degradation observed exclusively in the mouth areas of reconstructed frames confirms precise mask localization.}
    \label{fig:mask_vis}
\end{figure}

To validate our mouth-area localization strategy, we perform zero-masking experiments in the VAE latent space. As illustrated in Fig.~\ref{fig:lip_mask}, we first detect facial keypoints to localize the mouth region with a bounding box, map this box to VAE latent space considering spatial downsampling and temporal merging, set the corresponding latent features to zero, and finally decode back to pixel space. Accurate localization should result in degradation exclusively in the mouth area while preserving other regions.

Fig.~\ref{fig:mask_vis} presents validation results across four different video samples. For each sample, we compare the original frames with detected mouth boxes (top row) against the reconstructed frames from zero-masked latents (bottom row). As shown in the reconstructed sequences, the mouth regions exhibit clear visual degradation (blurring or distortion), while other facial features such as eyes, nose, and overall face structure remain well-preserved. This selective degradation pattern across different subjects, poses, and expressions confirms that our spatial-temporal downsampling approach accurately preserves mouth localization when mapping from pixel space to VAE latent space. The precise alignment validates that the mouth-area loss applies targeted supervision to the correct region for effective lip-speech synchronization learning.

\section{Future Work and Scaling}
\label{sec:future_work}

JoVA demonstrates strong performance across various joint video-audio generation and editing tasks, and its unified architecture opens several promising directions for further scaling and deployment.

Our current model is trained on a curated dataset of approximately 3 million video-audio-text pairs, which is sufficient to establish robust multimodal alignment and high-quality generation within the unified architecture. Scaling the training data to a significantly higher order of magnitude (e.g., tens or hundreds of millions of highly aligned pairs) could further unlock the potential of native joint self-attention. Exploring the scaling behavior and emergent capabilities of JoVA under such data regimes remains an important direction for future work.

\end{document}